\documentclass[a4paper,11pt]{article} 
\usepackage[osf,sc]{mathpazo}
\usepackage[scaled=0.90]{helvet}
\usepackage[scaled=0.85]{beramono}
\usepackage[utf8]{inputenc}
\usepackage[T1]{fontenc}
\usepackage[width=17cm,height=25cm]{geometry}

\usepackage{amsmath,amssymb,amsfonts}
\usepackage{algorithmic}
\usepackage{multirow}
\usepackage{graphicx}
\usepackage{textcomp}
\usepackage{xcolor}
\usepackage{tikz}
\usepackage{standalone}
\usepackage{tikz}
\usepackage{import}
\usetikzlibrary{calc,patterns,decorations.pathmorphing,decorations.markings,positioning,backgrounds,arrows,shapes,fit,matrix,spy,shapes.geometric}
\usepackage{pgfplots}
\pgfplotsset{width=10cm,compat=1.9}

\usepackage{balance}

\renewcommand{\vec}[1]{\boldsymbol{#1}}
\newcommand{\mat}[1]{\mathbf{#1}}
\renewcommand{\epsilon}{\varepsilon}
\newcommand{\vers}[1]{\overline{\vec{#1}}}

\usepackage[margin=1cm]{caption}
\def\R{\mathbb{R}}

\newcommand{\frm}[1]{\langle #1\rangle}

\begin{document}
    { 
        \noindent \itshape 
        XXIV IMEKO World Congress ``Think Metrology'' \\
    }
    { 
        \noindent \itshape 
        August 26 - 29, 2024, Hamburg, Germany \\ 
    }
    
    { 
        \noindent \huge \bfseries 
        Surface defect identification using \\
        Bayesian filtering on a 3D mesh \\
    }

    {
        \noindent \renewcommand\thefootnote{}%
        \noindent%
        Matteo Dalle Vedove$^{1,2}$,
        $\ $Matteo Bonetto$^{1}$,
        $\ $Edoardo Lamon$^{3,4}$,
        $\ $ Luigi Palopoli$^{3}$, \\
        $\ $Matteo Saveriano$^{1}$,
        $\ $Daniele Fontanelli$^{1}$%
        \footnotemark{}
        \footnotetext{\\
        $^{1}$Department of Industrial Engineering, Universit\`a di Trento, Trento, Italy. {\tt\small matteo.dallevedove@unitn.it} \\
            $^{2}$DRIM, Ph.D. of national interest in Robotics and Intelligent Machines. \\
            $^{3}$Department of Information Engineering and Computer Science, Universit\`a di Trento, Trento, Italy. \\
            $^{4}$Human-Robot Interfaces and Interaction, Istituto Italiano di Tecnologia, Genoa, Italy.
        }
    }

% \title{
%     \textit{ \normalsize XXIV IMEKO World Congress ``Think Metrology''\\ \vspace{-6mm}
%     August 26 - 29, 2024, Hamburg, Germany}\\ \vspace{2mm}
%     {\LARGE \bf Surface defect identification using Bayesian filtering on a 3D mesh}
%     \vspace{2mm}
%     {\large
%         Matteo Dalle Vedove$^{a,b,*}$,
%         Matteo Bonetto$^{a}$,
%         Edoardo Lamon$^{c,d}$,
%         Luigi Palopoli$^{c}$, \\
%         Matteo Saveriano$^{a}$,
%         Daniele Fontanelli$^{a}$ \\
%     }
%     \vspace{11pt}
%     { \normalsize
%     {$^{a}$ Department of
%       Industrial Engineering, Universit\`a di Trento, Trento,
%       Italy.}  \\
%     {$^{b}$ DRIM, Ph.D. of national interest in Robotics and
%       Intelligent Machines.}\\
%     {$^{c}$ Department of Information
%       Engineering and Computer Science, Universit\`a di Trento,
%       Trento, Italy.}\\
%     {$^{d}$ Human-Robot Interfaces and
%       Interaction, Istituto Italiano di Tecnologia, Genoa, Italy.} \\
%     {$^{*}$ Corresponding author: \tt\small matteo.dallevedove@unitn.it} \\
%          }
% }

\makeatletter

%% --- Authors
% \author{
%
%     \thanks{Co-funded by the European Union. Views and opinions
%       expressed are however those of the author(s) only and do not
%       necessarily reflect those of the European Union or the European
%       Commission. Neither the European Union nor the granting
%       authority can be held responsible for them. EU - HE Magician –
%       Grant Agreement 101120731.}  \thanks{$^{1}$Department of
%       Industrial Engineering, Universit\`a di Trento, Trento,
%       Italy. \tt\small matteo.dallevedove@unitn.it}
%     \thanks{$^{2}$DRIM, Ph.D. of national interest in Robotics and
%       Intelligent Machines.}  \thanks{$^{3}$Department of Information
%       Engineering and Computer Science, Universit\`a di Trento,
%       Trento, Italy.}  \thanks{$^{4}$Human-Robot Interfaces and
%       Interaction, Istituto Italiano di Tecnologia, Genoa, Italy.}
%   } % end of author block

% \author{}
% \maketitle

% \vspace{-5cm}

\begin{abstract}
    \noindent
    This paper presents a CAD-based approach for automated surface defect
detection. We leverage the a-priori knowledge embedded in a CAD model
and integrate it with point cloud data acquired from commercially
available stereo and depth cameras. The proposed method first
transforms the CAD model into a high-density polygonal mesh, where
each vertex represents a state variable in 3D space. Subsequently, a
weighted least squares algorithm is employed to iteratively estimate
the state of the scanned workpiece based on the captured point cloud
measurements. This framework offers the potential to incorporate
information from diverse sensors into the CAD domain, facilitating a
more comprehensive analysis. Preliminary results demonstrate promising
performance, with the algorithm achieving convergence to a
sub-millimeter standard deviation in the region of interest using only
approximately 50 point cloud samples. This highlights the potential of
utilising commercially available stereo cameras for high-precision
quality control applications.

\end{abstract}

% \begin{IEEEkeywords}
% component, formatting, style, styling, insert
% \end{IEEEkeywords}

\section{Introduction}
\label{sec:introduction}
Industry 4.0 relies on the circular dependency between the data collected
from the process and their exploitation to adapt and improve the
production rate and variety~\cite{Catalucci2022}.  In this context,
automatic defect detection in industrial production and assembly lines
is paramount for delivering products that match the continuously
increasing levels of quality requirements. A paradigmatic example in the automotive industry is the identification of small
defects on metallic surfaces that are the results of faulty welding operations (weld
splatters) or other mechanical processes (e.g., dents during the press operation).

According to current industrial practices, the detection of this type
of anomalies is performed by human operators, who are specially
trained to identify tiny defects in short amounts of time, dictated by
the requirements of the production process~\cite{Andersson2009}. In
the automotive industry, the takt time, i.e., the time allotted to
scan a whole car body surface, is usually in the order of some
minutes. At the same time, the task is repetitive and not particularly
engaging, hence it appears convenient to involve also robotic
operators in some of the steps required in the product quality checks
and rework. While it is possible to exploit advanced predictive
quality model to speed up the process~\cite{salcedo2020predicting},
still the task is non-trivial and presents challenging complexity
especially in the defect detection
phase. %predictive quality model of automotive paint shop
% The replacement of humans with robotic devices is a task of challenging complexity.
Promising results come from deflectometry, an
optical-based method which measures the deformation patterns of
structured light~\cite{ARNAL2017306,MOLINA2017263,ZHOU2020}.
%As observed by Sarosi et al.~\cite{Sarosi2010}, a proper application of deflectometry in this context requires a high reflectance of the inspected surface.
However, a proper application of deflectometry in this context requires a high reflectance of the inspected surface~\cite{sarosi2010detection}.
The direct implication of this requirement in the automotive industry is that vehicles can be inspected with this method only after the painting process.
Nevertheless, detecting a defect too late in the process comes with very high reworking costs, which are difficult to accept considering that most of these defects could be revealed right after weld and assembly processes, during the so-called body-in-white phase.

A different approach is to detect defects from images leveraging
machine learning methods.  Such methods have been extensively
investigated in the literature~\cite{Zhou2019, Xie2019,Zhang2020},
with deep-learning being predominant in recent
years~\cite{chang2020lightweight,Gao2020,Watson2020,block2021inspection,zhao2024attention}.
Leveraging high quality images in a heavily structured environment,
vision systems proved to be effective, with detection accuracy that
can exceed $95\%$ for some specific datasets.  The main limitation of
these techniques, however, resides in the fact that, to properly train
classifiers, big datasets are required and, therefore, human labour to
manually label all the possible defects is strictly needed. In
addition, they still lack of generalisation capabilities, hence, in
case of changes in the production line or for new products, new
labelled data is required.  Even though unsupervised learning methods
overcome this problem by autonomously learning patterns from the
data, %their exploitation in defect identification is still far,
they are still not mature nor effective enough due to the low accuracy
they usually provide~\cite{Lehr2021}.
% - \cite{zhao2024attention} deep learning fabric defect detection;
% - \cite{block2021inspection} automatic detection and classification of imprint defects on the surface of metal parts with convolutional neural networks
%Machine learning based systems threat the classification problem in a black-box approach, thus the prior and information-rich knowledge provided by CAD models, which are available in every industrial scenario, is poorly exploited.
On the other hand, a rich source of information that is always available is the CAD model of the workpiece.
This data is mainly used in geometric inspection~\cite{Germani2010, Sjodahl_2021} to assess whether components comply, or not, with the tolerances specified in the CAD model, but little research focused on the exploitation of 3D models to determine surface imperfections~\cite{long2023reconstruction}. %defect analysis with computed tomography on 3D CAD models

% - \cite{zhao2024attention} deep learning fabric defect detection;
% - \cite{long2023reconstruction} defect analysis with computed tomography on 3D CAD models
% - \cite{block2021inspection} automatic detection and classification of imprint defects on the surface of metal parts with convolutional neural networks
% - \cite{salcedo2020predicting} predictive quality model of automotive paint shop

% Defects are identified in images leveraging machine learning methods
% that, despite their encouraging performance in structured
% environments~\cite{Chang2019}, lack generalisation to different
% products and scenarios, limiting their exploitation in flexible
% production processes~\cite{Lehr2021}. On the other hand, a rich source
% of information that is always available is the CAD model of the
% workpiece.  This knowledge is mainly used for geometric tolerance
% inspection~\cite{Sjodahl_2021}, while little research compared the CAD
% with 3D point cloud measurements to classify different
% defects~\cite{Doring2006}.

In this paper, we address the surface defect detection problem by
combining the a-priori CAD model knowledge with several point-cloud
measurements acquired through commercially available stereo cameras. %a research avenue followed by a surprisingly small number of authors to the dates~\cite{Doring2006, long2023reconstruction}.
In our method, the CAD drawing is first translated into a high-density polygonal mesh, assigning a degree-of-freedom (a state variable) to each mesh vertex.
Based on this representation, we use a weighted least squares to recursively estimate the state of the scanned workpiece.
Additionally, this enables the embedding in the CAD domain of  information potentially coming from multiple kind of sensors.
Preliminary results shows that, with approximately 50 point-cloud samples, the algorithm converges to a state with sub-millimetric standard deviation in the region of interest, effectively showing the potential of commercial stereo and depth cameras in the context of quality inspection. %that usually requires high-end sensors.

% Once the surface state is estimated, defect segmentation is carried
% out by measuring the spatial correlation between close vertices.
% In the final setup, the camera will be mounted on a robotic manipulator to obtain full coverage of the workpiece.

% In this paper, we address the surface defect detection problem by
% combining the a-priori CAD model knowledge with several point-cloud
% measurements acquired through commercially available stereo cameras.
% The CAD is encoded by a high-density polygonal mesh.  We assign to
% each vertex a degree of freedom that encodes the surface state.  In
% virtue of this representation, the weighted least squares estimator
% fuses point cloud measurements in order to effectively reconstruct the
% workpiece mesh starting from the prior knowledge provided by the CAD.
% Once the surface state is estimated, defect segmentation is carried
% out by measuring spatial correlation between close vertices.
% In the final setup, the camera will be mounted on a robotic manipulator in order to provide full coverage of the workpiece.

This paper is organised like this. Sec.~\ref{sec:methods} reports some background knowledge and reports the proposed algorithm description, while
Sec.~\ref{sec:results} report the experimental setup and the results obtained. Finally, in Sec.~\ref{sec:conclusions}, we summarise our work and highlight future works.

\section{CAD-based Defect Detection Algorithm}
\label{sec:methods}

The solution for defect detection here presented builds upon the data
acquired from a stereo camera and the prior CAD description of the
component to check, whose availability is a customary assumption for
production lines. Therefore, this section at first presents the models
of the adopted instruments and of the assumed prior data, then we
describe how those information are optimally fused together using
Bayesian filtering techniques.

\subsection{CAD Model}
\begin{figure}[t]
  \centering \begin{tikzpicture}[scale=0.8]

    \coordinate (P1) at (0, 0);
    \coordinate (P2) at (3, 0);
    \coordinate (P3) at (6, -1);
    \coordinate (P4) at (5, 2);
    \coordinate (P5) at (2, 2);

    \coordinate (F1) at (1.75, 0.8);
    \coordinate (F2) at (3.2, 1.3);
    \coordinate (F3) at (4.6, 0.4);

    \draw (P1) -- (P2) -- (P5) -- cycle;
    \draw (P2) -- (P4) -- (P5) -- cycle;
    \draw (P2) -- (P3) -- (P4) -- cycle;

    \node[below left]   at (P1) {$\vec V_1$};
    \node[below]        at (P2) {$\vec V_2$};
    \node[below right]  at (P3) {$\vec V_3$};
    \node[above right]  at (P4) {$\vec V_4$};
    \node[above left]   at (P5) {$\vec V_5$};

    \node at (F1) {$F_1$};
    \node at (F2) {$F_2$};
    \node at (F3) {$F_3$};

\end{tikzpicture}
  \caption{Example of polygonal mesh with $n_v = 5$ vertices and
    $n_f = 3$ faces.}
  \label{fig:meshexample}
\end{figure}
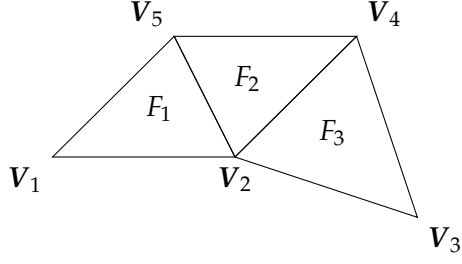

The main goal of the algorithm proposed in this paper is to match the point cloud collected from the sensors with the  3D model of the workpiece. The starting point of the algorithm is the stereolitography (STL) 3D model of the piece. The STL model can be easily generated by any CAD software, and it models each object as a set of triangles forming a polygonal mesh. 
Internally, the file stores a matrix $\mat V \in \mathbb R^{3\times n_v}$ of $n_v$ vertices in the space, and $\mat F \in \mathbb N^{3\times n_f}$ that encodes the vertices association of the $n_f$ faces constituting the polygonal mesh.
As an example, for the planar mesh in Fig.\ref{fig:meshexample}, the corresponding matrices $\mat V, \mat F$ are given by:
\begin{equation}
  \label{eq:VandF}
  \mat V = \begin{bmatrix}
    0 & 3 & 6 & 5 & 2 \\
    0 & 0 & -1 & 2 & 2 \\
    0 & 0 & 0 & 0 & 0
  \end{bmatrix} , \qquad \mat F = \begin{bmatrix}
    1 & 2 & 2 \\
    2 & 4 & 3 \\
    5 & 5 & 4
  \end{bmatrix},
\end{equation}
where $\vec V_i \in \mathbb R^3$ are the Euclidean coordinates of the
$i$-th vertex, corresponding to the $i$-th column of $\mat V$, while
the indexes of the vertices for the $j$-th face $F_j$ are listed in
the $j$-th column of $\mat F$. 
%-%
%-%
In the following, we denote by $\mathcal M$ a mesh, which is defined
by the vertices $\mat V$ and the faces $\mat F$.

\subsection{Point clouds}
Stereo cameras leverage the disparity map of 2 images taken at known
relative positions to associate a depth value with each pixel, thus
constructing a 3D representation of the scene, thus generating a point
cloud.  More in depth, given the $k$-th point cloud acquisition, a
point cloud can be seen as a set $\mathcal Z_k$ of $n_p$ 3D points in
the space described in the camera's reference frame $\frm{C}$:
\begin{equation}
  \mathcal Z_k = \left\{ \vec z_{k, i}^{(c)} \in \R^{3}, \ i = 1, \dots, n_p \right\}.
\end{equation}
From a metrological perspective, each registered element of the point
$\vec z_{k,i}^{(c)}$ is a nonlinear function $\vec h^\star$ (i.e., the
sensor model) of the actual scene in view $\mathcal S$ and of the
camera pose $\vec p_c \in SE(3)$ (where $SE(3)$ is the special
Euclidean group, that is $\R^3\times SO(3)\subset\R^6$), with the
addition of an uncertainty term $\vec \epsilon_{k,i} \in \R^3$
deriving from the camera and the reconstruction algorithm:
\begin{equation}
  \label{eq:measure}
  \vec z_{k,i}^{(c)} = \vec h^\star \left( \mathcal S, \vec p_c \right) + \vec \epsilon_{k,i}.
\end{equation}

An analytical derivation of $\vec\varepsilon_{k,i}$ is very difficult
since the uncertainties depend on the intrinsic and extrinsic
parameters of the camera, on the parameters of the built-in
point-cloud reconstruction algorithm and on the environmental
conditions.  To overcome this limitation, Ortiz et al. carried out in
\cite{Ortiz2018} a Type A statistical analysis on a commercial
stereo-camera
%, a StereoLabs ZED2 
to reconstruct the root-mean-square error (RMSE) of each collected
point.  Based on this work and in light of the Central Limit Theorem,
we assume $\vec\epsilon_{k,i} \sim \mathcal N(0, \mat R_{k,i})$ and
white, that is normally distributed, with zero mean, and with a
covariance matrix defined as
\begin{equation} \label{eq:covariance}
    \mat R_{k,i}(\rho) = a \, e^{b \rho} \mat I_{3\times3},
\end{equation}
with $a, b\in \mathbb R$ coefficients that depend on the image
acquisition resolution and $\rho \in \mathbb R^+$ the distance between
the camera origin and the acquisition point.

\subsection{Measurement model}

To define the measurement model, we define a state that captures the
dissimilarity between the inspected object and its reference shape
resulting from the CAD model. We first notice that the scene in view
$\mathcal{S}$ in~\eqref{eq:measure} corresponds to the mesh
$\mathcal{M}$, except for the presence of the defects. Indeed, by the
defining with $\mathcal{S}_n$ the scene for a perfect nominal
workpiece, the knowledge of $\vec p_c$ (given by an external
measurement system or by mounting the stereo camera on a robot end
effector) allows us to write
\begin{equation}
  \label{eq:Nominal}
  \vec h^{\star\star} \left( \mathcal{M} \right) = \vec h^{\star} \left( \mathcal{S}_n, \vec p_c \right) .
\end{equation}
Notice that the uncertainties in the knowledge of $\vec p_c$ are
embedded in $\epsilon_{k,i}$ in~\eqref{eq:measure}. To detect the
defect, we are interested in determining the regions of the mesh
$\mathcal{M}$ in which the equation in~\eqref{eq:Nominal} does not
hold. To this end, for each measurement $\vec z_{k,i}^{(c)}$
in~\eqref{eq:measure}, we first compute the point
$\overline{\vec z}_{k,i,j} \in \mathbb R^3$ obtained through a
ray-cast with the mesh $\mathcal M$, i.e.,
$\overline{\vec z}_{k,i,j} \in \mathcal{M}$ is the closest point to
$\vec z_{k,i}^{(c)}$ and belongs to the face $F_j$ of the
mesh. Therefore, $\overline{\vec z}_{k,i,j} = f(\vec
z_{k,i}^{(c)})$. For the nominal scene $\mathcal{S}_n$ and in the
ideal case of no uncertainties, there exists $j\in \{1,\dots,n_f\}$
such that $\vec z_{k,i}^{(c)} = \overline{\vec z}_{k,i,j}$. In the
presence of defects but still in the ideal case of no uncertainties,
we can model the difference using the dissimilarity measure
$x_{k,i,j}$, i.e.
$\vec h^{\star} \left( \mathcal{S}, \vec p_c \right)$
in~\eqref{eq:measure} becomes
\begin{equation}
  \label{eq:measfuncupdate}
  \vec h^{\star} \left(
    \mathcal{S}, \vec p_c \right) = \overline{\vec z}_{k,i,j} + x_{k,i,j}
  \vers n_j,
\end{equation}
where $\vers n_j$ is the normal direction of the $j$-th face, defined
using $\mat V$ and $\mat F$ in~\eqref{eq:VandF}. Hence, for each
gathered point cloud, we have an indirect measurement of $x_{k,i,j}$
for the $j$-th face, thus leading to the following indirect
measurement function
\begin{equation}
  \label{eq:measureMod}
  \vec \delta_{k,i,j} = g(\vec z_{k,i}^{(c)}) = \vec z_{k,i}^{(c)} -
  \overline{\vec z}_{k,i,j} = x_{k,i,j} \vers n_j.
\end{equation}
When the uncertainties $\vec \epsilon_{k,i}$ comes into play, we have
$f(\vec z_{k,i}^{(c)}) = \overline{\vec z}_{k,i,j} + \vec
\eta_{k,i,j}$, where $\vec \eta_{k,i,j}$ is the effect of
$\vec \epsilon_{k,i}$ through the ray tracing function. Therefore,
\eqref{eq:measureMod} turns to
\begin{equation}
  \label{eq:measureFinal}
  \vec \delta_{k,i,j} = x_{k,i,j}
  \vers n_j + \vec \epsilon_{k,i} + \vec \eta_{k,i,j} \approx x_{k,i,j}
  \vers n_j + \vec \epsilon_{k,i} ,
\end{equation}
where we have assumed that, being $\vec \eta_{k,i,j}$ a function of
$\vec \epsilon_{k,i}$ and using the actual mesh as reference, the
effect of $\vec \eta_{k,i,j}$ is already embedded into
$\vec \epsilon_{k,i}$. By noticing that all the cloud points falling
on the face $F_j$ are just multiple measurement of the same quantity
$x_{k,j}$, we simply define the state to be estimated as
$\vec x_k = [x_{k,1}, \dots, x_{k,n_f}]^\top\in \R^{n_f}$, which
quantifies the deviation between the nominal model and the
measurements: whenever a state component approaches zero, we expect a
low discrepancy between the corresponding face of the two meshes. On
the other hand, a high value of $|x_{k,i}|$ is reasonably associated
with a high discrepancy and hence with the presence of a defect in the
area.

\subsection{Bayesian filtering}

In order to build an estimator for the vector $\vec x_k$, we start
with a weighted least squares solution. First, we define the vector of
measurements~\eqref{eq:measureFinal} as
\[
  \vec \Delta_k = \begin{bmatrix} \vec \delta_{k,1,j_1}^\top & \vec
    \delta_{k,2,j_2}^\top & \dots & \vec \delta_{k,n_p,j_{n_p}}^\top
    \end{bmatrix} ,
\]
where $j_l$ is the index of the face associated with the $l$-th
measurement. Therefore
\[
  \vec \Delta_k = \mat H_k \vec x_k ,
\]
where $\mat H_k \in \mathbb R^{3n_p \times 3 n_f}$. In particular
\[
  \vec \delta_{k,i,j_{i}}^\top = \mat H_k^{(i)} \vec x_k ,
\]
where the $3\times 3n_f$ block $\mat H_k^{(i)}$ of $\mat H_k$
pertaining to the $i$-th indirect measurement is a matrix of all zeros
except for the $j_i$-th column, which is equal to $\vers
n_j$. Given~\eqref{eq:measureFinal}, we can also immediately derive
that the covariance matrix of the measurement uncertainties of
$\Delta_k$ is given by
\begin{equation}
  \label{eq:meascovariance}
  \mat R_k =
  \textrm{blkdiag}\left\{ \mat R_{k,i}\big( \|\vec z_{k,i}^{(c)} \|\big), \ i
    = 1,\dots, n_p \right\} \in \R^{3n_p\times 3n_p} ,
\end{equation}
where $\rho = \|\vec z_{k,i}^{(c)} \|$ in~\eqref{eq:covariance} has
been used and where the point cloud uncertainties are approximated as
uncorrelated in space. A more detailed analysis on this specific point
is left for future works.

% Our strategy to address these issues if by a numeric approximation, both of the function $\vec h_k \in \mathbb R^{3n_f}$ and of its Jacobian $\mat H_k \in \mathbb R^{3n_p \times 3 n_f}$.
% Firstly, in~\eqref{eq:wlsupdate}, the measurement vector $\vec z_k \in \mathbb R^{3n_p}$ is constructed  by stacking each registered point $\vec z_{k,i} \in \mathbb R^3$ for the current timestamp $k$, i.e.,
% \begin{equation} \label{eq:measurement}
%     \vec z_k = \big( \vec z_{k,1}^\top, \vec z_{k,2}^\top, \dots, \vec z_{k,n_p}^\top\big)^\top,
% \end{equation}
% where each measurement has been projected from the camera reference frame to the common system defined by the CAD model.
% By construction, measurement noise $\vec \epsilon$ is normally distributed with zero mean and covariance

% Since the function $\vec h_k$ represents the expected measurement given the current estimated state $\vec x_k$, we construct it as follows.
% Denoted with $\overline{\vec z}_{k,i} \in \mathbb R^3$ the point obtained through a ray-cast with the mesh $\mathcal M$ starting from the camera origin and directed as $\vec z_{k,i}$, then the 3-dimensional block $\vec h_k^{(i)}$ associated to the $i$-th measured point is set as
% \begin{equation} \label{eq:measfuncupdate}
%     \vec h_k^{(i)} = \overline{\vec z}_{k,i} + x_{k,j} \vers n_j,
% \end{equation}
% with $j$ being the index of the triangular face the ray was cast on.

At the beginning of the algorithm, we initialise our estimates with
two dummy values $\hat{\vec x}_0$ and $\vec P_0$: $\hat{\vec x}_0$ is
the first estimate of the actual $\vec x_k = \vec x$ (since the
defects did not change along the point clouds collection); $\vec P_0$
is the rated covariance matrix of the estimation error
$\vec x - \hat{\vec x}_0$. Then, a recursive weighted least-squares
(RWLS) estimation algorithm, i.e., a Bayesian filter, can be applied,
thus having for $k \geq 1$
\begin{equation} \label{eq:wlsupdate}
\begin{aligned}
  \mat S_{k+1} & = \mat H_{k+1} \mat P_k \mat H_{k+1}^\top + \mat
  R_{k+1} , \\
  \mat W_{k+1} & = \mat P_k \mat H_{k+1}^\top \mat S_{k+1}^{-1} , \\
  \hat{\vec x}_{k+1} & = \hat {\vec x}_k + \mat W_{k+1} \big( \vec
  \Delta_{k+1} -
  \mat H_{k+1} \hat{\vec x}_{k+1} \big) , \\
  \mat P_{k+1} & = \left(\mat I - \mat W_{k+1} \mat H_{k+1} \right)
  \mat P_k .
\end{aligned}
\end{equation}

\subsection{Information filter}
From a numerical point of view, using a RWLS estimator might not be
beneficial due to the high dimensionality of the problem.  In fact,
each point-cloud usually comprises tens of thousands of points, while
the STL derived from a complex CAD might easily reach millions of
faces. Since with the application of the RWLS we are implicitly
considering that the state
$\vec x \sim \mathcal N(\hat{\vec x}_k, \mat P_k)$ is drawn from a
Gaussian distribution, we can reduce the numerical complexity
resorting to the dual representation of~\eqref{eq:wlsupdate} in the
information domain.  Let $\vec \xi_k \in \R^{3n_f}$ and
$\mat \Omega_k \in \R^{3n_f\times 3n_f}$ be respectively the
information vector and matrix, given the transformation
\begin{equation}
  \label{eq:coordchange}
  \mat P_k = \mat \Omega_k^{-1}
  \mbox{ and } \hat{\vec x}_k = \mat \Omega_k^{-1} \vec \xi_k,
\end{equation}
then~\eqref{eq:wlsupdate} is mapped into
\begin{equation} \label{eq:informationfilter}
\begin{aligned}
    \mat \Omega_{k+1} & = \mat H_k^\top \mat R_k^{-1} \mat H_k + \mat \Omega_k \\
    \vec \xi_{k+1} & = \mat H_k^\top \mat R_k^{-1} \vec z_k + \vec \xi_k.
\end{aligned}
\end{equation}
This algorithm requires less computation, thus increasing the
computational efficiency. In addition, information representations are
sparse, thus code can be optimised to reduce memory requirements and
computational costs~\cite{probrobots}.

\section{Experimental Analysis}
\label{sec:results}

\begin{figure}[t]
  \centering \import{Figures/}{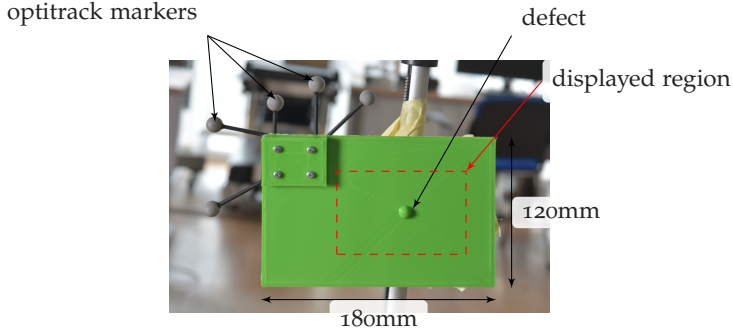}
  \caption{3D printed tablet containing one defect in the middle of
    the board.  The area bounded by the red line is the region that
    will be displayed in the following figures.}
  \label{fig:tablet}
\end{figure}

To evaluate the proposed algorithm, we designed and 3D printed a flat
tablet, shown in Fig.~\ref{fig:tablet}, presenting in the middle a
unique spherical defect of radius $5$~mm.
Doing so, we have a ground truth to address the properties of the
estimation algorithm.  Instead, the performance of the detection
algorithm have been computed considering the defect-free shape of the
CAD, i.e., a flat rectangular surface.  With this set-up, the
algorithm can be tested against different polygonal densities of the
mesh, so as to assess the perception performance of the system in
different settings.  In the following, the mesh size represents the
minimum edge length of the right-angled equilateral triangle resulting
from the meshing algorithm.

Measurements have been collected from 2 commercially available stereo
cameras: the Intel RealSense D415, and the StereoLabs Zed2. Both
cameras were used at the resolution $1\,280\times 720$px and,
according to~\cite{Ortiz2018}, we select in~\eqref{eq:covariance}
$a=0.0184$ and $b=0.2106$.  The RealSense camera is an active sensor,
so, to reduce the uncertainty of the point-cloud computation, it
exploits a projected infra-red pattern on the environment.  On the
other hand, the Zed2 is a passive sensor, thus it requires a good
illumination to improve the quality of the reconstructed
point-cloud. For this reason, we illuminated the tablet sample with 2
soft-boxes photo lights.

To provide he camera pose $\vec p_c$ in~\eqref{eq:measure}, we used
the OptiTrack motion capture system.  Since the provided data were not
of sufficient to match the CAD model with the registered point-clouds
with millimetre precision, at each iteration the method calls an
iterative closest point (ICP) algorithm to improve alignement of the
measurement with the known CAD model.  The algorithm has been
implemented in C++, using ROS2 as middleware to communicate between
cameras and the OptiTrack system.

\subsection{Results}

For each of the following tests, the algorithm is initialised with
zero state $\hat{\vec x}_0 = 0$ and a diagonal covariance
$\mat P_0 = \sigma^2_0 \mat I$, with $\sigma_0 = 50$~mm.  To evaluate
the algorithm and compare the two camera systems, we use the RMSE
defined as
\begin{equation} \label{eq:rmse} \textrm{RMSE}(k) =
  \sqrt{\frac{1}{n_{f,v}} \vec e_k^\top \vec e_k }, \qquad \vec e_k =
  \big(\hat{\vec x}_k - \vec x\big) \mat \Pi,
\end{equation}
with $\hat{\vec x}_k$ the state estimate at the $k$-th iteration
(i.e., when $k$ point clouds are acquired), $\vec x$ is the reference
state that is obtained by geometrical difference between the CAD model
of the defective and defect-free objects, and
$(\pi_{ij}) = \mat \Pi \in \R^{n_f\times n_{f,v}}, \pi_{ij} \in
\{0,1\}$ is a matrix used to select the $n_{f,v}$ triangular faces
involved in the measurements.  In fact, only half of the tablet is
directly visible from the camera, thus only those polygons can be
updated.  In addition, since our method relies on ray-casting to
create correspondences with the mesh, areas close to the edges appear
to be highly distorted, so all triangle within $6$~mm from the border
have been neglected in the performance metrics.

\begin{figure}[t]
  \centering \includegraphics[width=0.55\linewidth]{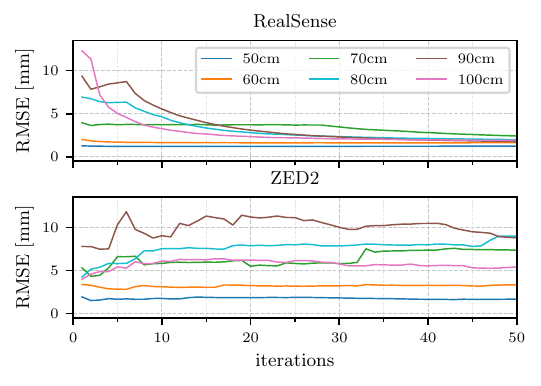}
  \caption{RMSE of the algorithm computed on a $5$~mm as function of
    the iterations and relative distances between object and camera
    for the two compared devices.}
  % Python source: rmse-zed-realsense-distance.py
  \label{fig:rmse-distance}
\end{figure}

As first step, we compared the algorithm outcomes with the two cameras while inspecting the tablet at different distances, with the tablets normal pointing toward the camera.
As shown in Fig.~\ref{fig:rmse-distance}, it is possible to observe that the RealSense camera, even at different acquisition distances, converges to a RMSE slightly above the $2$~mm mark.
%-%
%-%
On the other hand, the Zed2 shows a trend for which, as distance
increases, the RMSE increases as well.

\begin{figure*}[t]
  \centering
  \begin{tabular}{cc}
    \includegraphics[width=0.48\linewidth]{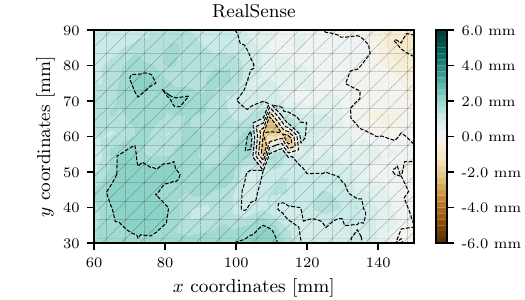}
    & \includegraphics[width=0.48\linewidth]{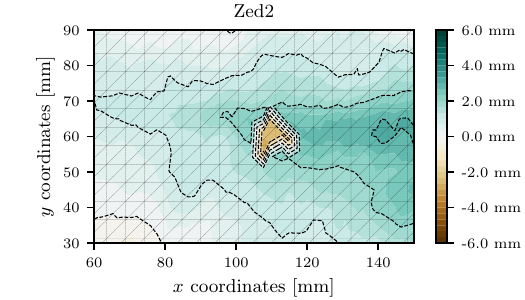} \\
    (a)
    &(b)
  \end{tabular}
  \caption{Estimation error
    $\vec e_{50} = \hat{\vec x}_{50} - \vec x$, in the region of interest, for the
    Realsense (a) and Zed2 (b) cameras at the $50$-th
    iteration. Measurements have been taken at $50$~cm, and the mesh
    has size of $5$~mm. Drawn isolines mark the barrier at each whole
    integer step of error. The reported triangular grid is the one of
    the mesh actually used to carry out the estimation. }
  \label{fig:state-error}
  % Python source: error_plots.py
  % then copy into Figures/state-map-<>.pdf the desired figure present
  % in Figures/Source/Plots/ErrorMap/*
\end{figure*}

To push forward the analysis, we report the distribution of the state
estimation error on the testing tablet. As shown in
Fig.~\ref{fig:state-error}, the state estimate provided by the
RealSense camera is qualitatively better, appearing consistent with
the reality, while the Zed2 present several issues.
%-%
%-%
This is the case notwithstanding the different light conditions:
irrespective of the adoption of natural environment illumination,
positioning one soft-box on one side of the tablet, and positioning 2
soft-boxes at both sides, the estimates we obtained were comparable.

These results can be directly linked with the different camera working
principles.  Being the RealSense an active sensor, the projected light
pattern on the tablet enables a more accurate and dense point-cloud
reconstruction, while, at a relatively low distance from the object,
the passive Zed2 camera struggles in the identification of the
correspondences using the disparity map only.  In numbers, each
triangular face of the $5$~mm mesh in the region of interest is
sampled throughout the $50$ iterations, on average, $2\,200$ times
using the RealSense, and $150$ times using the Zed2.  This is also
reflected on the standard deviation of the estimated state that, for
tests shown in Fig.~\ref{fig:state-error}, are respectively $0.25$~mm
and $0.95$~mm for the RealSense and the Zed2.
% For all this reasons, from now onward analysis will be carried out mainly for the RealSense camera.

\begin{figure}[t]
  \centering
  \includegraphics[width=0.55\linewidth]{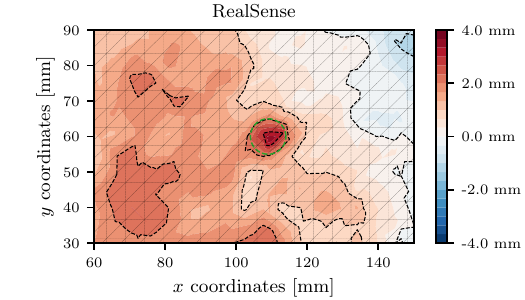}
  \caption{State estimate $\hat{\vec x}_{50}$, in the region of interest, for the
    Realsense camera and location of the spherical defect (dashed
    green line). Experiment setup as in Fig.~\ref{fig:state-error}.The
    reported triangular grid is the one of the mesh actually used to
    carry out the estimation. }
    \label{fig:state-map}
    % Python source: error_plots.py
    % then copy into Figures/state-map-<>.pdf the desired figure
    % present in Figures/Source/Plots/ErrorMap/*
\end{figure}

By looking at Fig.~\ref{fig:state-map}, which depicts the state
estimate obtained by collecting $50$ measurements at a distance of
$50$~cm using the RealSense camera, it is clear that the system is
able to capture the presence of the defect.
%-%
%-%
Still, we must also note that outside of the region of interest, the
state estimate appears diverging.  Looking at different experimental
outcomes, it emerges a pattern for which the central region of the
tablet is overestimates, while peripheral areas are under-estimated.
These errors can be associated to a non-correct calibration of the
sensing devices. The measurement model indeed considers the
uncertainties $\vec \epsilon_{k,i}$ to have zero means, i.e. the
measurement has been compensated from any sistematic error.  Even
though the measurement uncertainty covariance~\eqref{eq:covariance}
proved to be effective in general, no model compensation is performed.
It is indeed possible that, at the distances we tested the camera,
optical reconstruction distortions can introduce a bias in the
measurements.  The deeper investigation of this issue is left for
future developments.

After multiple tests, we concluded that best results are obtained at
the minimum distance of $50$~cm, due to a lower measurement
uncertainty given by the model~\eqref{eq:covariance}, and a higher
amount of casted rays on the mesh.  At this distance, then we also
analysed the impact of the relative orientation between the camera
pointing direction and the tablet normal direction. As reported in
Table~\ref{tab:heading-comparison}, where the mean and the standard
deviation of the absolute estimation error is reported, the RealSense
camera confirms better performance in this situation as well.
%-%
\begin{table}[t]
  \centering
  \caption{Mean and standard deviation of the absolute error at the
    $50$-th iteration of the algorithm as function the relative
    heading between camera orientation and the tablet normal. All
    acquisition took at a distance of $50$~cm with mesh of $5$~mm.}
  \label{tab:heading-comparison}
  
          \begin{tabular}{c c c c c}
            
            \multirow{2}{*}{Angle} & \multicolumn{2}{c}{RealSense} &  \multicolumn{2}{c}{Zed2} \\
            & $\textrm{mean}(|\vec e_{50}|)$ & $\textrm{std}(|\vec e_{50}|)$ & $\textrm{mean}(|\vec e_{50}|)$ & $\textrm{std}(|\vec e_{50}|)$ \\
            \hline \hline
    
            0$^\circ$ & 1.02mm & 0.73 mm & 1.21mm & 1.13mm \\
10$^\circ$ & 0.84mm & 0.54 mm & 3.18mm & 3.16mm \\
15$^\circ$ & 0.84mm & 0.62 mm & 3.04mm & 2.40mm \\
20$^\circ$ & 0.80mm & 0.63 mm & 3.58mm & 3.37mm \\
30$^\circ$ & 0.95mm & 0.79 mm & 4.27mm & 3.59mm \\
45$^\circ$ & 0.83mm & 0.82 mm & 8.02mm & 10.73mm \\
60$^\circ$ & 1.04mm & 1.09 mm & 17.24mm & 17.72mm \\
75$^\circ$ & 1.49mm & 1.52 mm & 22.85mm & 19.64mm \\
          \end{tabular}
          
\end{table}
%-%
Within the $50^\circ$ relative heading, the RealSense appears to work
consistently to reality, and all tests clearly show the pattern
associated to the non-correct compensation of the measurement bias.
On the other hand, Zed2 camera results get worse as the heading
increases.

\begin{figure}[t]
  \centering
  \includegraphics[width=0.55\linewidth]{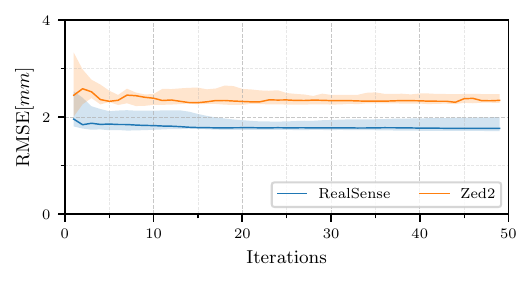}
  \caption{Median and quartiles of the RMSE distribution, as a
    function of the algorithm iterations, for the 2 cameras. For each
    camera, we performed $10$ tests at a distance of $50$~cm and zero
    relative heading. Tests carried out on mesh of size $5$~mm.
    }
    % Python source: rmse-zed-realsense-distance.py
  \label{fig:rmse-repeteability}
\end{figure}

Finally, to assess the repeatability of the algorithm, we performed
multiple trials with fixed experimental conditions.  In this case, as
depicted in Fig.~\ref{fig:rmse-repeteability}, tests carried out with
the RealSense camera shows a convincing convergence patterns.
%-%
%-%
After about $30$ iterations, the RMSE for the RealSense converges to a
characteristic value that can be related both to acquisition distance
and relative heading.  On the other hand, the algorithm executed on
the Zed2 camera has an inconsistent behavior, highlighting how the
sensor does not provide satisfactory results both in terms of accuracy
and repeatability.

One final, yet remarkable note, consists in the choice of the CAD mesh
size. By construction, the state represent the normal deviation
averaged on each triangle, thus to have a good spatial resolution,
smaller polygons are requested.  This, in turn, increases the number
$n_f$ of faces that, at some point, will make the RWLS
algorithm~\eqref{eq:wlsupdate}, even in it's information
representation~\eqref{eq:informationfilter}, unsolvable in reasonable
time.  In all these tests, we found that the $5$~mm mesh was the best
compromise in terms of spatial resolution, and computational time,
with point-cloud processing requiring $10$~s per algorithm iteration.
In practice, this solution as is, can't be used in a online system due
to the computational burden. However it is also true that, for the
sake of simplicity, code has been developed using a sequential
approach, but the algorithm formulation is heavily parallelisable on
dedicated hardware, so with ad-hoc implementation we expect a drastic
reduction of processing time.

\section{Conclusions}
\label{sec:conclusions}
 
In this work, we evaluated the use of the CAD knowledge of a workpiece
within a RWLS to incrementally estimate surface defects using
commercially available stereo-cameras.  Within the proposed approach,
the estimator converges to a standard deviation of $0.3$~mm in
approximately $40$~s, i.e., in $1$~s of point-cloud acquisitions.
While we acknowledge the limitations of the approach, potential
solutions are within reach. Calibration issues can be addressed
through bias modelling techniques. To overcome computational
limitations, the algorithm can be parallelised for more efficient
processing on modern hardware.  Furthermore, the current
implementation focuses on static measurements. Exploiting the motion
provided by robotic manipulators, future works will investigate
planning strategies to optimise information gathering based on the
specific workpiece geometry and potential defect types. This will
allow for a more targeted and efficient inspection process.  Overall,
the proposed method paves the way for utilising commercially available
stereo cameras for high-precision quality control applications. By
leveraging CAD models and addressing the identified limitations, this
approach has the potential to become a robust and cost-effective
solution for automated surface defect detection in various
manufacturing and inspection scenarios.

% Point that can be touched:
% \begin{itemize}
%     \item calibration issue $\Rightarrow$ we can do bias modelling
%     \item computational limitation $\Rightarrow$ parallelise algorithm
%     \item now we consider static measurements $\Rightarrow$ choose planning strategy to optimise information.
% \end{itemize}
% \input{Sections/appendix.tex}

\section*{ACKNOWLEDGEMENTS}
Co-funded by the European Union. Views and opinions
expressed are however those of the author(s) only and do not
necessarily reflect those of the European Union or the European
Commission. Neither the European Union nor the granting
authority can be held responsible for them. EU - HE Magician –
Grant Agreement 101120731.

% \printbibliography
\balance
\bibliographystyle{IEEEtran}
\bibliography{references}

\end{document}